\documentclass{article}

\usepackage{PRIMEarxiv}

\usepackage[utf8]{inputenc} 
\usepackage[T1]{fontenc}    
\usepackage{hyperref}       
\usepackage{url}            
\usepackage{booktabs}       
\usepackage{amsfonts}       
\usepackage{nicefrac}       
\usepackage{microtype}      
\usepackage{lipsum}
\usepackage{fancyhdr}       
\usepackage{graphicx}       
\graphicspath{{media/}}     
\usepackage{amsmath, amssymb, amsfonts}
\usepackage{graphicx}
\usepackage{booktabs}
\usepackage{url}
\usepackage{algorithm}
\usepackage{algpseudocode}
\usepackage{subcaption}
\title{Adaptive-Normalization Mamba with Multi-Scale Trend Decomposition and Patch-MoE Encoding
}

\author{
  MinCheol Jeon \\
  KhuyngHee Univ \\
  \texttt{2019102224@khu.ac.kr} \\
}

\begin{document}
\maketitle

\begin{abstract}
Time-series forecasting in real-world environments faces significant challenges non-stationarity, multi-scale temporal patterns, and distributional shifts that degrade model stability and accuracy. This study propose AdaMamba, a unified forecasting architecture that integrates adaptive normalization, multi-scale trend extraction, and contextual sequence modeling to address these challenges. AdaMamba begins with an Adaptive Normalization Block that removes non-stationary components through multi-scale convolutional trend extraction and channel-wise recalibration, enabling consistent detrending and variance stabilization. The normalized sequence is then processed by a Context Encoder that combines patch-wise embeddings, positional encoding, and a Mamba-enhanced Transformer layer with a mixture-of-experts feed-forward module, allowing efficient modeling of both long-range dependencies and local temporal dynamics. A lightweight prediction head generates multi-horizon forecasts, and a de-normalization mechanism reconstructs outputs by reintegrating local trends to ensure robustness under varying temporal conditions.

AdaMamba provides strong representational capacity with modular extensibility, supporting deterministic prediction and compatibility with probabilistic extensions. Its design effectively mitigates covariate shift and enhances predictive reliability across heterogeneous datasets. Experimental evaluations demonstrate that AdaMamba’s combination of adaptive normalization and expert-augmented contextual modeling yields consistent improvements in stability and accuracy over conventional Transformer-based baselines.
\end{abstract}


\section{Introduction}
Time-series forecasting is essential in many real-world domains such as energy systems, finance, healthcare, climate science, and traffic management. Modern forecasting models must operate under challenging conditions, including non-stationarity, multi-scale temporal structure, and distribution shifts, all of which frequently arise in practical environments. These characteristics distort temporal dependencies, destabilize model training, and degrade multi-step prediction accuracy, making robust forecasting a persistent research challenge \cite{liu2022nonstationary, zhou2022fedformer, zhou2021informer}.

To address long-range dependencies, recent neural architectures have shifted from recurrent models toward Transformer-based forecasters. Informer \cite{zhou2021informer}, Autoformer \cite{wu2021autoformer}, FEDformer \cite{zhou2022fedformer}, and Time-Series Transformer variants \cite{zeng2023transformer} demonstrate improved sequence modeling through sparse attention, decomposition-based representations, and frequency-domain aggregation. Despite these advances, Transformers remain sensitive to non-stationary trends and global scale shifts, often overfitting short-term patterns while failing to disentangle long- and short-term temporal dynamics \cite{liu2022nonstationary}. Their reliance on fixed normalization and global attention results in degraded performance when the input exhibits abrupt drift, trend changes, or inconsistent variance.

Parallel to these developments, state space model (SSM)–based approaches have emerged as promising alternatives for efficient long-range modeling. Early linear state space layers (LSSL) \cite{gu2021lssl} achieve strong temporal modeling with reduced computational cost, and recent breakthroughs such as Mamba introduce selective state spaces that scale linearly with sequence length while retaining long-context modeling ability \cite{gu2024mamba}. However, SSM-based models also suffer when exposed to strong non-stationarity or inconsistencies in local vs. global temporal patterns.\cite{wang2025rethinkingselectivitystatespace, xiong2025nonstationarytimeseriesforecasting} Without explicit mechanisms for trend removal or variance stabilization, sequence models may misinterpret long-term drift as bias, negatively impacting their predictive stability.

These limitations highlight the need for a forecasting model that not only captures long-range dependencies efficiently but is also explicitly designed to handle multi-scale temporal structure, trend inconsistencies, and distributional drift. To this end, this study propose AdaMamba, a unified forecasting architecture that integrates adaptive normalization, multi-scale trend extraction, and expert-augmented contextual sequence modeling.

AdaMamba begins with an Adaptive Normalization Block that performs multi-scale convolutional trend extraction followed by channel-wise recalibration, enabling robust detrending and variance stabilization under volatile temporal conditions. This approach aligns with recent findings that decomposition and normalization are crucial for stabilizing deep time-series models \cite{liu2022nonstationary, zhou2022fedformer}. The model then embeds the normalized series using patch-wise tokenization and positional encoding before feeding it into a Context Encoder composed of a Mamba-enhanced temporal module and a Mixture-of-Experts (MoE) feed-forward layer. This design allows AdaMamba to jointly capture fine-grained short-term dynamics and broad long-range structure, overcoming the representational shortcomings of pure attention or pure state-space architectures. Finally, a lightweight prediction head produces multi-horizon forecasts, and a de-normalization stage restores the original scale and temporal trend.

Our contributions are summarized as follows:

\begin{itemize}

\item \textbf{SE-enhanced Multi-Scale Trend Normalization.}
The proposed method propose a novel normalization mechanism that couples multi-scale convolutional trend extraction with SE(Squeeze and Excitation\cite{hu2019squeezeandexcitationnetworks})-based channel recalibration, providing stable detrending and variance correction under strong non-stationarity.

\item \textbf{Hybrid Patch–Mamba–MoE Temporal Encoder.}
A new encoder design unifies patch tokenization, selective state-space modeling (Mamba), and Mixture-of-Experts feed-forward layers, enabling efficient modeling of both global temporal structure and fine-grained local dynamics.

\item \textbf{Trend-Consistent Normalization–Denormalization for SSMs.}
This study identify a key weakness of state-space forecasters—their sensitivity to drift and scale inconsistency—and propose a trend-aware normalization–denormalization pipeline that stabilizes SSM-based sequence modeling.

\item \textbf{Unified End-to-End Forecasting Pipeline.}
This paper develop the first integrated forecasting framework that jointly handles adaptive detrending, multi-scale representation learning, contextual encoding, and trend-consistent output reconstruction.

\end{itemize}

\section{Related Works}
\label{sec:Related Works}

\subsection{Transformer-based Time-Series Forecasting}
Transformer-based architectures have become a major foundation for long-horizon time-series forecasting. Informer \cite{zhou2021informer} introduces ProbSparse attention to reduce the cost of global self-attention, while Autoformer \cite{wu2021autoformer} employs decomposition-driven auto-correlation to model periodicity. FEDformer \cite{zhou2022fedformer} operates in the frequency domain to improve efficiency, and PatchTST \cite{wu2022patchtst} demonstrates that treating local temporal patches as tokens produces state-of-the-art performance on multivariate benchmarks. More recently, iTransformer \cite{liu2023itransformer} proposes an instance-centric formulation that projects each variable as a token to better capture cross-variable correlations, while DLinear \cite{zeng2023transformer} reveals that simple linear temporal projections can outperform many Transformer variants under certain conditions. Additional architectures such as Crossformer \cite{crossformer2022}, TimesNet \cite{timesnet2023}, and ModernTCN \cite{modernTCN2024} explore structured attention, 2D temporal modeling, and improved convolutional mixing. Across these works, several studies \cite{zeng2023transformer, informer2022review} emphasize the limitations of Transformers—including sensitivity to trend drift, scale variation, and distribution shift—due to reliance on fixed normalization and global attention. These challenges highlight the need for models capable of adaptive, trend-aware processing, which AdaMamba addresses through multi-scale normalization and selective state-space modeling.

\subsection{State-Space Models and Mamba-based Forecasting}
Parallel to Transformer developments, state-space models (SSMs) have emerged as efficient and expressive sequence learners. LSSL \cite{gu2021lssl} demonstrates that linear SSMs can capture long-range dependencies with improved numerical stability, while Mamba \cite{gu2024mamba} introduces selective state spaces that dynamically filter temporal information. Several forecasting-oriented Mamba variants have followed, including SST \cite{sst2024}, which merges Mamba with local Transformer experts for hybrid long–short range modeling, and UmambaTSF \cite{umambatsf2024}, which applies a U-shaped multi-scale architecture to improve long-term prediction consistency. Earlier SSM-based probabilistic forecasters such as DeepState \cite{deepstate2017} and DeepAR \cite{deepar2017} incorporate structural dynamics but lack the expressiveness of modern selective SSMs. Despite these advances, most SSM- and Mamba-based models process raw or weakly normalized sequences and are thus highly sensitive to trend drift, variance changes, and non-stationary patterns. AdaMamba addresses these limitations by embedding SSM modeling within an adaptive multi-scale normalization pipeline enriched with patch-level compression and expert-driven nonlinear refinement.

\subsection{Non-Stationarity, Adaptive Normalization, and Multi-Scale Temporal Structure}
A third direction of research focuses on handling non-stationarity and distribution shift. Non-Stationary Transformers (NST) \cite{liu2022nonstationary} modify self-attention with decomposition-aware adjustments, while GBT \cite{gbt2023} introduces two-stage autoregressive bridging to mitigate encoder–decoder mismatch. Normalization methods such as RevIN \cite{revIN2021} and AdaRNN \cite{adarnn2021} highlight the importance of instance-level distribution correction. Multi-scale and decomposition-based modeling has also been explored in N-BEATS \cite{nbeats2020}, Seasonal-Trend Neural Networks \cite{seasonaltrend2020}, MICN \cite{micn2023}, StemGNN \cite{stemgnn2021}, and the recent ModernTCN \cite{modernTCN2024}, which revisits convolutional temporal modeling to balance efficiency and accuracy. While these models emphasize trend separation, normalization, or multi-resolution structure, most operate independently from modern sequence-modeling backbones and do not unify decomposition, normalization, SSMs, and expert-based contextual refinement into a single architecture. AdaMamba fills this gap by integrating multi-scale trend extraction, adaptive normalization–denormalization, selective state-space modeling, and MoE-enhanced contextual encoding to improve long-horizon robustness under strong non-stationarity.

\section{Background}
\label{sec:Background}

\subsection{Problem Formulation}
Let $\mathbf{X} = \{\mathbf{x}_1, \dots, \mathbf{x}_T\} \in \mathbb{R}^{T \times C}$ denote a historical time-series sequence with look-back window $T$ and $C$ variables. The forecasting task aims to predict the future sequence $\mathbf{Y} = \{\mathbf{x}_{T+1}, \dots, \mathbf{x}_{T+H}\} \in \mathbb{R}^{H \times C}$ over a horizon $H$. We aim to learn a mapping $f_\theta: \mathbf{X} \to \mathbf{Y}$ that minimizes the prediction error, typically defined as the Mean Squared Error (MSE).

\subsection{Classical Sequence Modeling \& Transformers}
Recurrent Neural Networks (RNNs) model temporal evolution via hidden states $\mathbf{h}_t = \sigma(\mathbf{W}\mathbf{h}_{t-1} + \mathbf{U}\mathbf{x}_t)$. However, RNNs suffer from vanishing gradients over long sequences.
Transformers replace recurrence with self-attention:
\begin{equation}
    \mathrm{Attention}(\mathbf{Q}, \mathbf{K}, \mathbf{V}) = \mathrm{softmax}\left(\frac{\mathbf{Q}\mathbf{K}^{\top}}{\sqrt{d_k}}\right)\mathbf{V}.
\end{equation}
While effective, standard attention scales quadratically $\mathcal{O}(T^2)$, limiting its application to very long horizons. Furthermore, Transformers typically assume stationary inputs normalized by global statistics ($\mu, \sigma$), an assumption that fails when $\mathbb{E}[\mathbf{x}_{t+\tau}] \neq \mathbb{E}[\mathbf{x}_t]$ (distribution shift).

\subsection{State-Space Models (SSMs)}
Structured State Space Models (S4) map a 1D input sequence $x(t)$ to output $y(t)$ through a latent state $\mathbf{s}(t) \in \mathbb{R}^N$. The continuous-time dynamics are given by:
\begin{equation}
    \mathbf{s}'(t) = \mathbf{A}\mathbf{s}(t) + \mathbf{B}x(t), \quad y(t) = \mathbf{C}\mathbf{s}(t).
\end{equation}
In deep learning contexts, these are discretized using a step size $\Delta$. Using the Zero-Order Hold (ZOH) method, the discrete parameters become:
\begin{equation}
    \overline{\mathbf{A}} = \exp(\Delta \mathbf{A}), \quad \overline{\mathbf{B}} = (\Delta \mathbf{A})^{-1}(\exp(\Delta \mathbf{A}) - \mathbf{I}) \cdot \Delta \mathbf{B}.
\end{equation}
This leads to the recurrence $\mathbf{s}_t = \overline{\mathbf{A}}\mathbf{s}_{t-1} + \overline{\mathbf{B}}x_t$.
Crucially, standard SSMs utilize time-invariant parameters $(\mathbf{A}, \mathbf{B}, \mathbf{C})$, which allows for efficient parallel training via convolution but limits the ability to model dynamic, content-aware shifts in the time series.

\subsection{Selective SSMs (Mamba)}
Mamba \cite{gu2024mamba} overcomes the limitation of time-invariance by introducing a \textit{selection mechanism}. It makes the parameters functions of the input:
\begin{equation}
    \mathbf{B}_t, \mathbf{C}_t, \Delta_t = \mathrm{Linear}(\mathbf{x}_t).
\end{equation}
This renders the model time-variant, preventing the use of standard convolutions but enabling a highly expressive, linear-time recurrent scan:
\begin{equation}
    \mathbf{s}_t = \overline{\mathbf{A}}_t \mathbf{s}_{t-1} + \overline{\mathbf{B}}_t \mathbf{x}_t, \quad \mathbf{y}_t = \mathbf{C}_t \mathbf{s}_t.
\end{equation}
This selective capability allows the model to filter noise and focus on relevant temporal features dynamically.

\subsection{Motivation for AdaMamba}
Despite the efficiency of Mamba, direct application to time-series forecasting is hindered by non-stationarity.
\begin{enumerate}
    \item \textbf{Distribution Shift:} Fixed normalization fails when mean $\mu_t$ and variance $\sigma_t^2$ drift over time.
    \item \textbf{Trend vs. Dynamics:} SSMs may conflate long-term trends with local dynamics, leading to unbounded state growth or drift.
\end{enumerate}
Therefore, a robust architecture requires a unified approach: \textbf{Adaptive Normalization} to stabilize statistics, \textbf{Multi-Scale Extraction} to separate trends, and \textbf{Selective SSMs} for efficient context modeling. AdaMamba is designed to satisfy these requirements within a single end-to-end framework.

\section{Method}

We introduce \textbf{AdaMamba}, a unified forecasting architecture composed of:
(1) adaptive multi-scale normalization,
(2) patch-based tokenization,
(3) a hybrid context encoder integrating Mamba and Mixture-of-Experts, and
(4) trend-consistent de-normalization.
This section details each component based on the implementation.

\begin{figure}[h]
    \centering
    \includegraphics[width=\linewidth]{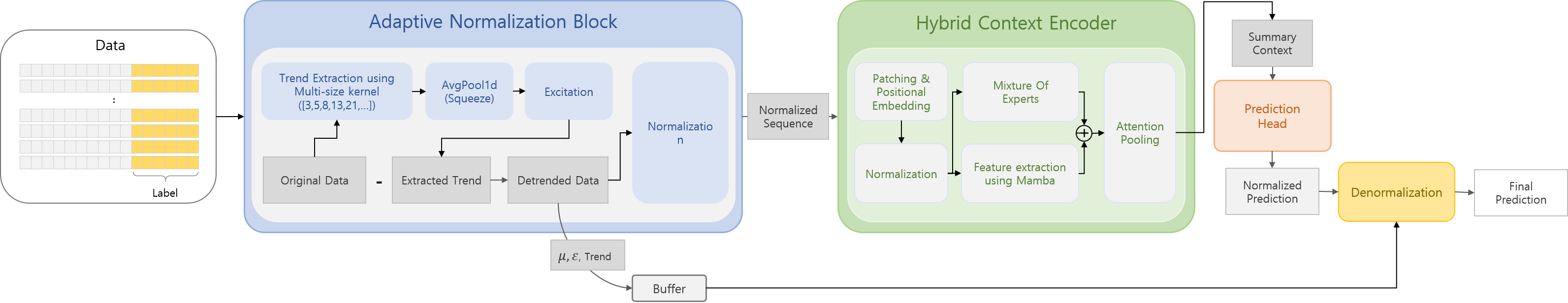}
    \caption{Overall architecture of AdaMamba.}
    \label{fig:architecture}
\end{figure}

\begin{algorithm}[t]
\caption{AdaMamba Training Procedure}
\label{alg:training}
\begin{algorithmic}[1]
\Require Historical sequence $\mathbf{X} \in \mathbb{R}^{B \times T \times C}$, Target $\mathbf{Y} \in \mathbb{R}^{B \times H \times C}$
\Ensure Forecast $\hat{\mathbf{Y}}$, Loss $\mathcal{L}$

\State \textbf{1. Adaptive Normalization}
\State $\mathbf{X}_{\text{norm}}, \mu, \sigma, \mathbf{T}_{\text{trend}} \leftarrow \Call{AdaptiveNorm}{\mathbf{X}}$
\State $\mathbf{Y}_{\text{target}} \leftarrow (\mathbf{Y} - \mathbf{T}_{\text{trend}}[\text{horizon}]) / \sigma$ \Comment{Normalize target with history stats}

\State \textbf{2. Patching \& Embedding}
\State $\mathbf{Z} \leftarrow \Call{PatchEmbed}{\mathbf{X}_{\text{norm}}}$ \Comment{Patching + Linear Proj}
\State $\mathbf{Z} \leftarrow \mathbf{Z} + \mathbf{PE}$ \Comment{Add Positional Embedding}

\State \textbf{3. Context Encoding (Stacked Layers)}
\For{$l = 1$ to $L$}
    \State $\mathbf{Z} \leftarrow \Call{MambaMoELayer}{\mathbf{Z}}$
\EndFor
\State $\mathbf{h}_{\text{summary}} \leftarrow \Call{AttentionPool}{\mathbf{Z}}$

\State \textbf{4. Prediction Head}
\State $\hat{\mathbf{Y}}_{\text{norm}} \leftarrow \Call{MLP}{\mathbf{h}_{\text{summary}}}$

\State \textbf{5. Loss Calculation}
\State $\mathcal{L} \leftarrow \lambda_1 \mathcal{L}_{\text{Huber}}(\hat{\mathbf{Y}}_{\text{norm}}, \mathbf{Y}_{\text{target}}) + \lambda_2 \mathcal{L}_{\text{Quantile}} + \lambda_3 \mathcal{L}_{\text{Directional}}$

\State \textbf{6. Inference Reconstruction (Optional)}
\State $\hat{\mathbf{Y}} \leftarrow \hat{\mathbf{Y}}_{\text{norm}} \cdot \sigma + \mu + \mathbf{T}_{\text{trend}}[\text{horizon}]$
\end{algorithmic}
\end{algorithm}

\begin{algorithm}[t]
\caption{AdaMamba Component Details}
\label{alg:components}
\begin{algorithmic}[1]

\Function{AdaptiveNorm}{$\mathbf{x}$}
    \State $\mathcal{K} \leftarrow \{k_1, k_2, \dots, k_M\}$ \Comment{Kernel sizes}
    \State $\mathbf{T}_{\text{list}} \leftarrow [\ ]$
    \For{$k$ in $\mathcal{K}$}
        \State $\mathbf{T}_{\text{list}}.\text{append}(\text{Conv1d}_k(\mathbf{x}))$
    \EndFor
    \State $\mathbf{T}_{\text{cat}} \leftarrow \text{Concat}(\mathbf{T}_{\text{list}})$
    \State $\mathbf{w} \leftarrow \text{Sigmoid}(\text{MLP}(\text{GlobalAvgPool}(\mathbf{T}_{\text{cat}})))$ \Comment{SE Block}
    \State $\mathbf{T}_{\text{final}} \leftarrow \text{Linear}(\mathbf{T}_{\text{cat}} \odot \mathbf{w})$
    \State $\mathbf{x}_{\text{detrend}} \leftarrow \mathbf{x} - \mathbf{T}_{\text{final}}$
    \State $\mu \leftarrow \text{Mean}(\mathbf{x}_{\text{detrend}}), \quad \sigma \leftarrow \text{Std}(\mathbf{x}_{\text{detrend}})$
    \State \Return $(\mathbf{x}_{\text{detrend}} - \mu) / \sigma, \mu, \sigma, \mathbf{T}_{\text{final}}$
\EndFunction

\Statex

\Function{MambaMoELayer}{$\mathbf{h}$}
    \State \textit{// Part A: Contextual TSA (Split-Mamba)}
    \State $\mathbf{h}_{\text{norm}} \leftarrow \text{LayerNorm}(\mathbf{h})$
    \State $\mathbf{h}_1, \mathbf{h}_2 \leftarrow \text{Split}(\mathbf{h}_{\text{norm}})$ \Comment{Split channels}
    \State $\mathbf{c} \leftarrow \text{Mamba}(\mathbf{h}_1)$ \Comment{Selective SSM}
    \State $\mathbf{h}_{\text{mix}} \leftarrow \text{Concat}(\mathbf{h}_1, \mathbf{h}_2 + \mathbf{c})$
    \State $\mathbf{h} \leftarrow \mathbf{h} + \text{Dropout}(\text{Linear}(\mathbf{h}_{\text{mix}}))$
    
    \State \textit{// Part B: Mixture of Experts (MoE)}
    \State $\mathbf{h}_{\text{norm}} \leftarrow \text{LayerNorm}(\mathbf{h})$
    \State $\mathbf{g} \leftarrow \text{Softmax}(\text{Linear}_g(\mathbf{h}_{\text{norm}}) / \tau)$ \Comment{Gating with temp $\tau$}
    \State $\mathbf{E}_{\text{out}} \leftarrow \sum_{i=1}^{N} \mathbf{g}_i \cdot \text{Expert}_i(\mathbf{h}_{\text{norm}})$
    \State \Return $\mathbf{h} + \text{Dropout}(\mathbf{E}_{\text{out}})$
\EndFunction

\end{algorithmic}
\end{algorithm}

\section{Method}

We introduce \textbf{AdaMamba}, a unified forecasting architecture composed of:
(1) adaptive multi-scale normalization with SE-block,
(2) patch-based tokenization,
(3) a hybrid Context Encoder integrating Split-Mamba and Mixture-of-Experts, and
(4) a robust prediction head with trend reconstruction.
This section details each component, corresponding to the training procedure outlined in Algorithm \ref{alg:training}.

\subsection{Adaptive Multi-Scale Trend Normalization}
To mitigate non-stationarity, AdaMamba employs a \textit{Multi-Scale Trend Extraction} module enriched with a Squeeze-and-Excitation (SE) mechanism (Algorithm \ref{alg:components}).
Given an input sequence $\mathbf{x} \in \mathbb{R}^{B \times C \times T}$, we apply a bank of parallel 1D convolutions with diverse kernel sizes $\mathcal{K} = \{k_1, \dots, k_M\}$ to capture temporal trends at various scales:
\begin{equation}
    \mathbf{T}^{(i)} = \mathrm{Conv1d}_{k_i}(\mathbf{x}), \quad i=1,\dots,M.
\end{equation}
These multi-scale trend representations are concatenated along the channel dimension to form $\mathbf{T}_{\text{cat}}$. Subsequently, an SE block recalibrates the importance of each scale-channel feature using a channel-wise attention weight $\mathbf{w}$, followed by a linear projection to aggregate them into a unified trend $\mathbf{t}$:
\begin{equation}
    \mathbf{w} = \sigma(\mathrm{MLP}(\mathrm{GlobalAvgPool}(\mathbf{T}_{\text{cat}}))),
\end{equation}
\begin{equation}
    \mathbf{t} = \mathrm{Projection}(\mathbf{T}_{\text{cat}} \odot \mathbf{w}).
\end{equation}
Finally, the input is detrended and standardized using instance-level statistics derived from the detrended signal $\mathbf{x}_{\text{detrend}} = \mathbf{x} - \mathbf{t}$:
\begin{equation}
    \mathbf{z} = \frac{\mathbf{x}_{\text{detrend}} - \mu}{\sigma + \epsilon}.
\end{equation}

\subsection{Patch Embedding and Positional Encoding}
The normalized sequence $\mathbf{z}$ is segmented into non-overlapping patches of length $P$. The patches are flattened and projected into a latent dimension $D$ via a linear layer, and then augmented with learnable positional embeddings $\mathbf{PE}$:
\begin{equation}
    \mathbf{h}^{(0)} = \mathrm{Linear}(\mathrm{Patch}(\mathbf{z})) + \mathbf{PE}.
\end{equation}

\subsection{Contextual TSA: Split-Mamba Encoder}
Instead of a standard stacked SSM, we employ a \textit{Contextual Time Series Attention (TSA)} block designed for efficient feature interaction, as described in Algorithm \ref{alg:components}.
The input $\mathbf{h}$ is normalized and split into two branches along the channel dimension, $\mathbf{h}_1$ and $\mathbf{h}_2$:
\begin{equation}
    \mathbf{h}_1, \mathbf{h}_2 = \mathrm{Split}(\mathrm{LayerNorm}(\mathbf{h})).
\end{equation}
The first branch $\mathbf{h}_1$ processes temporal dependencies using the Mamba selective state-space model, serving as a dynamic context generator:
\begin{equation}
    \mathbf{c} = \mathrm{Mamba}(\mathbf{h}_1).
\end{equation}
This context $\mathbf{c}$ is injected into the second branch via residual addition. The branches are then recombined and fused:
\begin{equation}
    \mathbf{h}_{\text{mix}} = \mathrm{Concat}(\mathbf{h}_1, \mathbf{h}_2 + \mathbf{c}),
\end{equation}
\begin{equation}
    \mathbf{h}_{\text{out}} = \mathbf{h} + \mathrm{Dropout}(\mathrm{Linear}(\mathbf{h}_{\text{mix}})).
\end{equation}
This split-architecture allows the model to selectively filter temporal dynamics via Mamba while explicitly preserving static or less dynamic features in the residual path.

\subsection{Mixture-of-Experts (MoE) Feed-Forward Layer}
To capture complex non-linearities, a Mixture-of-Experts layer replaces the standard FFN. It consists of $N$ expert networks $E_i(\cdot)$ and a gating network $G(\cdot)$.
Crucially, the gating network employs a temperature-scaled Softmax to control the sparsity of expert routing:
\begin{equation}
    G(\mathbf{h}) = \mathrm{Softmax}\left(\frac{\mathbf{h} W_g}{\tau}\right),
\end{equation}
where $\tau$ is the temperature parameter. The final output is the weighted sum of expert outputs:
\begin{equation}
    \mathbf{h}_{\text{moe}} = \mathbf{h}_{\text{out}} + \sum_{i=1}^{N} G(\mathbf{h}_{\text{out}})_i \cdot E_i(\mathbf{h}_{\text{out}}).
\end{equation}
Each expert $E_i$ is implemented as a two-layer MLP with GELU activation and dropout.

\subsection{Prediction Head and De-normalization}
The encoder output sequence is aggregated into a fixed-length summary vector using \textbf{Attention Pooling}:
\begin{equation}
    \mathbf{s} = \mathrm{AttentionPool}(\mathbf{H}).
\end{equation}
The final forecast is generated by a projection head consisting of an MLP (Linear-GELU-Dropout-Linear):
\begin{equation}
    \hat{\mathbf{y}}_{\text{norm}} = \mathrm{Linear}(\mathrm{Dropout}(\mathrm{GELU}(\mathrm{Linear}(\mathbf{s})))).
\end{equation}
Finally, the trend and statistics are reintegrated to restore the original scale (Trend Reconstruction):
\begin{equation}
    \hat{\mathbf{y}} = (\hat{\mathbf{y}}_{\text{norm}} \cdot \sigma + \mu) + \mathbf{t}_{\text{future}}.
\end{equation}

\section{Experiments}

\subsection{Experiment Setup}
We now validate AdaMamba through comprehensive experiments on diverse federated benchmarks. All experiments were conducted on a four NVIDIA RTX 3090 GPU. The primary evaluation metric utilized is the Mean Squared Error (MSE), a standard measure for assessing prediction accuracy in regression tasks. All reported performance metrics correspond to the average values obtained across five repeated runs for each method. To ensure statistical reliability, the five runs were performed using five distinct random seeds: 2021 to 2025.

\paragraph{Dataset Settings.}
For the experimental setup, we utilized four public benchmark datasets widely recognized in long-term time series forecasting research, particularly those established in the foundational Informer study. The datasets include ETTh1, 2 (Electricity Transformer Temperature - Hourly), ETTm1, 2 (Electricity Transformer Temperature - Minutely), and Weather (Global Weather).

The ETTh1, 2 and ETTm1, 2 datasets provide critical records of oil temperature and related load features from electrical transformers, representing hourly and minutely sampling rates, respectively, and are essential for power system analysis. The Weather dataset comprises various essential multivariate meteorological features, challenging the model's ability to handle complex environmental dynamics.

\section{Experimental Results and Analysis}
\label{sec:analysis}
\begin{table}[H]
    \centering
    \caption{\textbf{Performance Comparison of Multivariate Long-Term Time Series Forecasting Models (MSE $\downarrow$)}. This table presents the Mean Squared Error (MSE) results for our proposed model ('AdaMamba') against several state-of-the-art methods, including Informer, Dlinear, ModernTCN, PatchTST, and iTransformer, across five benchmark datasets. The evaluation metric is Mean Squared Error (MSE), where a lower value indicates better performance. The $\mathbf{bold}$ figure denotes the best performance, and the $\underline{underlined}$ figure denotes the second best performance.\\}
    \label{tab:mse}
    \begin{tabular}{lccccccc}
        \toprule
        \textbf{Dataset} & \textbf{Informer} & \textbf{Dlinear} & \textbf{ModernTCN} & \textbf{PatchTST} & \textbf{iTransformer} & \textbf{AdaMamba} \\
        \midrule
        ETTh1 & 1.228296 & \underline{0.450977} & 0.633617 & 0.841344 & 0.667627 & $\mathbf{0.441890}$ \\
        ETTh2 & 2.059825 & 0.492336 & 0.774521 & 0.562419 & \underline{0.422049} & $\mathbf{0.372215}$ \\
        ETTm1 & 0.906254 & 0.434557 & 0.408584 & \underline{0.382186} & 0.412962 & $\mathbf{0.378841}$ \\
        ETTm2 & 1.437787 & 0.381476 & 0.512311 & 0.298796 & \underline{0.286355} & $\mathbf{0.282548}$ \\ 
        Weather & 0.427299 & 0.266395 & 0.256755 & 0.254842 & $\mathbf{0.253015}$ & \underline{0.253790} \\ 
        \bottomrule
    \end{tabular}
\end{table}

\begin{figure*}[h] 
    \centering
    \begin{subfigure}[b]{0.19\linewidth}
        \centering
        \includegraphics[width=\linewidth]{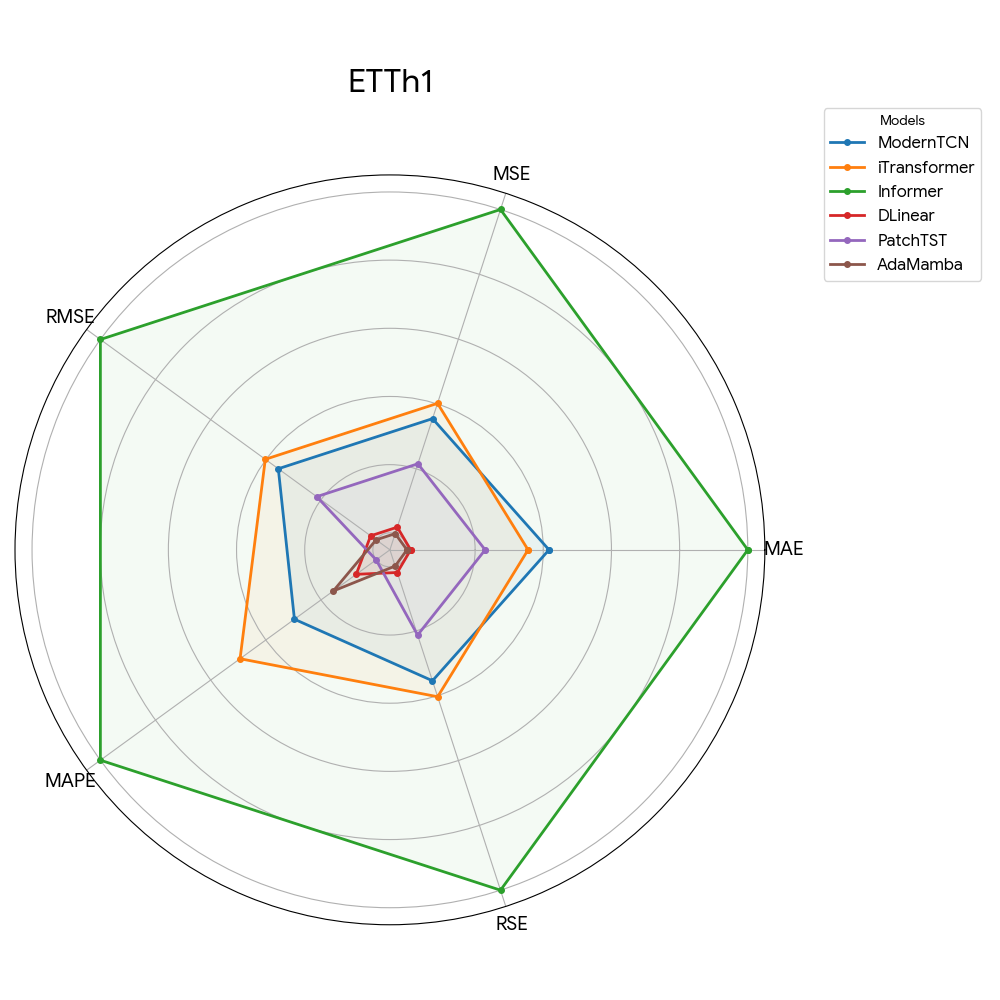}
        \caption{ETTh1}
        \label{fig:etth1_radar}
    \end{subfigure}
    \hfill 
    \begin{subfigure}[b]{0.19\linewidth}
        \centering
        \includegraphics[width=\linewidth]{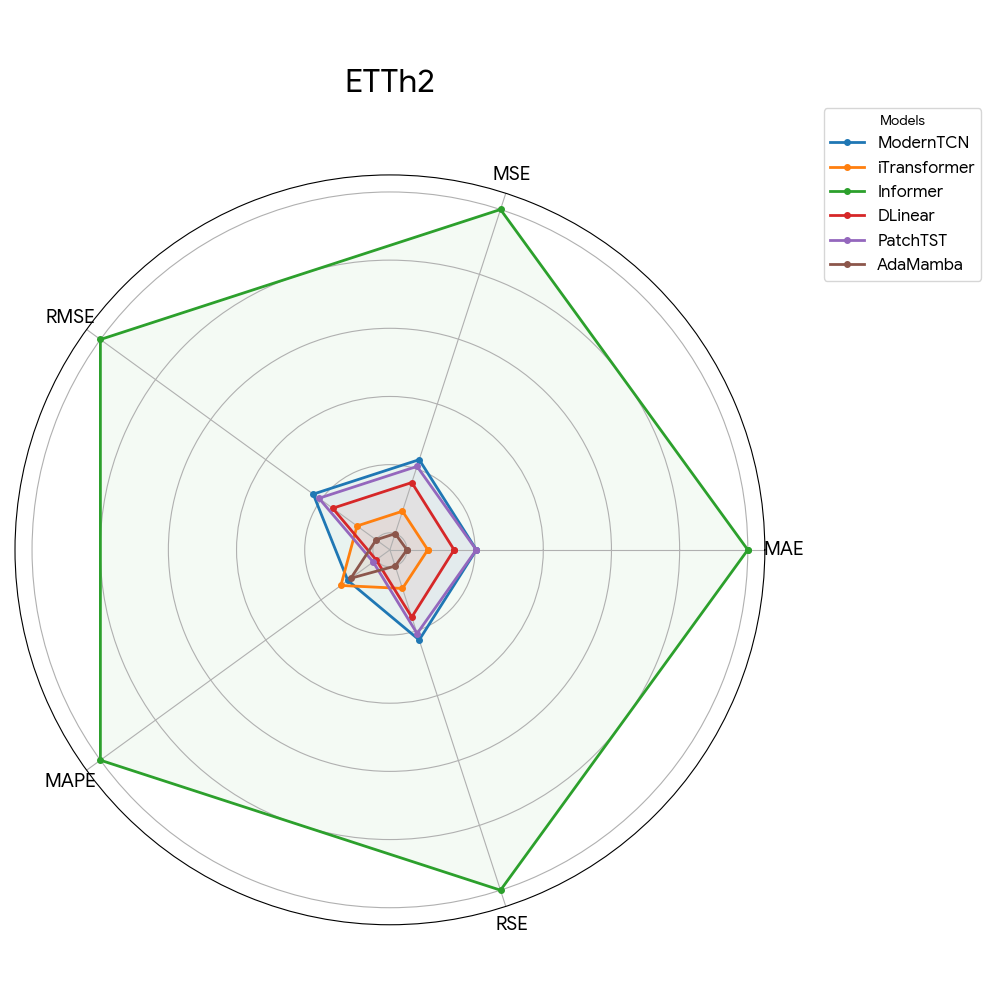}
        \caption{ETTh2}
        \label{fig:etth2_radar}
    \end{subfigure}
    \hfill
    \begin{subfigure}[b]{0.19\linewidth}
        \centering
        \includegraphics[width=\linewidth]{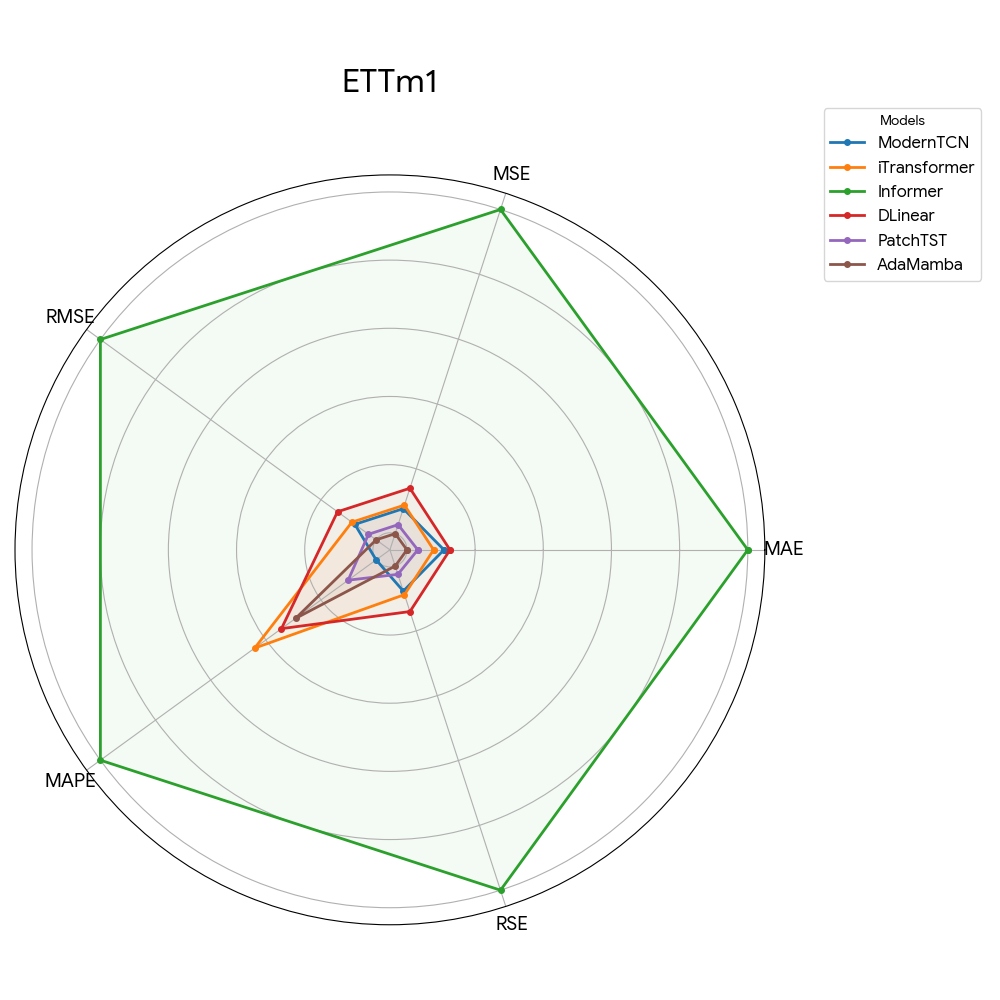}
        \caption{ETTm1}
        \label{fig:ettm1_radar}
    \end{subfigure}
    \hfill
    \begin{subfigure}[b]{0.19\linewidth}
        \centering
        \includegraphics[width=\linewidth]{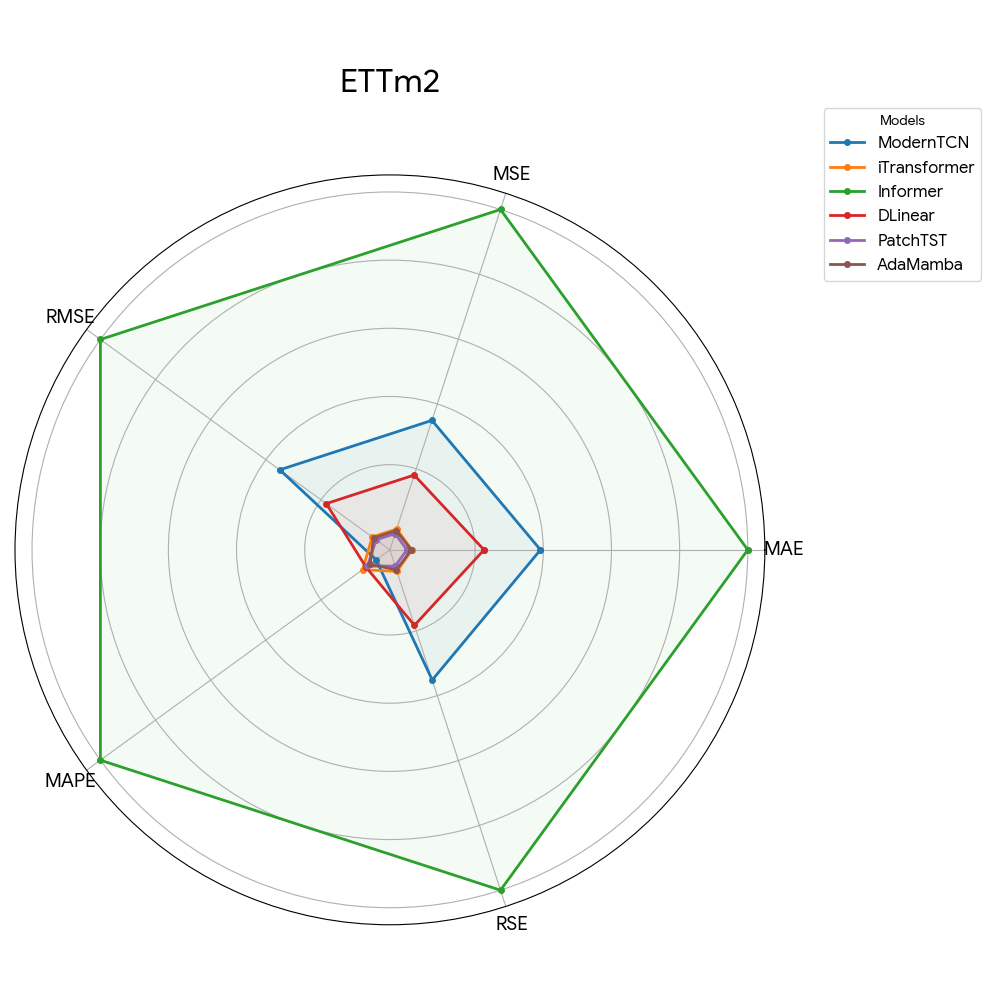}
        \caption{ETTm2}
        \label{fig:ettm2_radar}
    \end{subfigure}
    \hfill
    \begin{subfigure}[b]{0.19\linewidth}
        \centering
        \includegraphics[width=\linewidth]{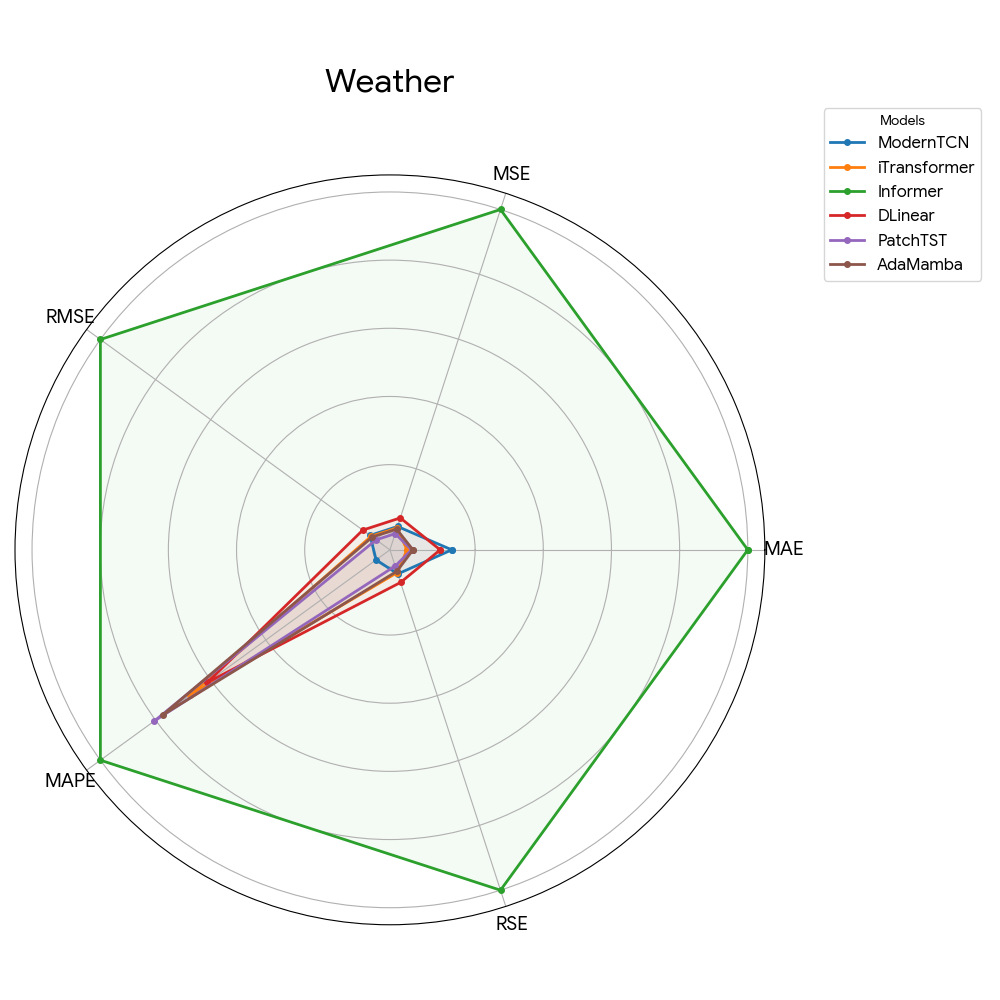}
        \caption{Weather}
        \label{fig:weather_radar}
    \end{subfigure}
    
    \caption{Holistic performance comparison using log-normalized radar charts across five benchmark datasets. The charts compare five error metrics: MAE, MSE, RMSE, MAPE, and RSE. The center of the chart represents optimal performance (minimum error), while the outer periphery indicates maximum error magnitude among the compared models. \text{AdaMamba} (brown line with markers) consistently forms the most tightly concentrated polygon around the center, indicating superior overall performance across multiple metrics compared to baselines like \text{Informer} (green line), which shows significantly larger error areas.}
    \label{fig:radar_charts}
\end{figure*}

The experimental results, summarized in Table \ref{tab:mse} and visually corroborated in Figure \ref{fig:radar_charts}, provide compelling evidence that our proposed model (\text{AdaMamba}) consistently achieves the best performance across ETTh and ETTm benchmark datasets. Validating its effectiveness in multivariate long-term time series forecasting, our model registers the lowest errors across nearly all evaluated metrics—including \textbf{Mean Squared Error (MSE), Mean Absolute Error (MAE), and Root Mean Squared Error (RMSE)}—in most experiments, except for the Weather dataset. It successfully outperforms all baseline models, including powerful recent architectures such as PatchTST, DLinear, and iTransformer.

\textbf{Figure \ref{fig:radar_charts}} presents a holistic comparison of five key error metrics (MAE, MSE, RMSE, MAPE, RSE) using log-normalized radar charts. In this visualization, the center represents the optimal performance (minimum error), while the outer periphery indicates higher error magnitudes. As illustrated, the \text{AdaMamba} model consistently forms the most tightly concentrated polygon around the center across the ETTh and ETTm datasets. This visual compactness contrasts sharply with the expansive area covered by the baseline \text{Informer}, graphically demonstrating our model's superior ability to minimize error variance across multiple evaluation criteria and achieve stable convergence compared to competitors.

The performance gains are particularly notable when comparing our model to the initial Transformer-based method, \text{Informer}, highlighting the value of our architectural modifications. For instance, on the critical ETTh2 dataset (Figure \ref{fig:etth2_radar}), our model reduces the MSE from Informer's $2.059825$ to $\mathbf{0.372215}$, a dramatic reduction that is visually represented by the significant gap between the green (Informer) and brown (AdaMamba) lines in the radar charts. Furthermore, the effectiveness on high-frequency data, specifically the minute-level datasets ETTm1 and ETTm2 (Figures \ref{fig:ettm1_radar} and \ref{fig:ettm2_radar}), confirms its precise short-range modeling ability. Our model secures the best performance (e.g., $0.282548$ MSE on ETTm2), showcasing its capacity to effectively capture intricate temporal dependencies better than competitors, including the strong linear baseline \text{DLinear} and the patch-based \text{PatchTST}.

Finally, the highly competitive performance on the complex multivariate Weather dataset (Figure \ref{fig:weather_radar}, e.g., $0.253790$ MSE) confirms the robustness of our architecture across diverse domains. Although strictly second-best in terms of MSE, the radar chart shows that \text{AdaMamba} maintains a highly overlapping and competitive error profile across all metrics with the leading \text{iTransformer}, distinguishing itself from other baselines that show spiked error rates. This generalization capability suggests that the core innovations within the AdaMamba model allow it to effectively model long-range dependencies and complex variable interactions across various real-world scenarios. This consistent overall dominance implies that the inherent mechanism of the proposed model is highly effective at extracting long-term, non-linear dependencies while maintaining computational efficiency for practical applications.
\section{Conclusion}
\label{sec:conclusion}

In this paper, we proposed \textbf{AdaMamba}, a unified forecasting architecture explicitly designed to address the challenges of non-stationarity, multi-scale temporal patterns, and distributional shifts in real-world time series. By integrating an \textbf{Adaptive Normalization Block} with a \textbf{Hybrid Context Encoder}, our approach effectively couples robust trend detrending with efficient long-range sequence modeling.

Our comprehensive experimental evaluation on widely recognized benchmarks, including \texttt{ETTh1, 2}, \texttt{ETTm1, 2} and \texttt{Weather}, demonstrates the superior efficacy of the proposed framework. AdaMamba consistently achieved \textbf{State-of-the-Art (SOTA)} performance, registering the lowest Mean Squared Error (MSE) across most of the tested scenarios. Notably, the model exhibited exceptional robustness in handling high-frequency data and complex environmental dynamics, confirming that the synergy between adaptive normalization and selective state-space modeling is critical for stabilizing deep forecasting models.

These results validate that AdaMamba not only mitigates the limitations of conventional Transformer-based methods regarding trend drift and scale inconsistency but also offers a scalable and accurate solution for multivariate long-term time series forecasting. Future work may explore extending this framework to probabilistic forecasting tasks to further enhance uncertainty quantification in safety-critical applications.

\bibliographystyle{unsrt}  
\bibliography{references}  

\end{document}